\documentclass{svproc}
\usepackage{url}
\usepackage{graphics} 
\usepackage{graphicx}
\usepackage{amsmath} 
\DeclareMathOperator*{\argmax}{arg\,max}

\begin{document}
\mainmatter              
\title{Learning Descriptor of Constrained Task from Demonstration }
\titlerunning{Descriptor of Constrained Task}  
%
\author{Xiang Zhang\inst{1} \and 
Matteo Saveriano\inst{1} \and Justus Piater\inst{1}}
\authorrunning{Xiang Zhang et al.} 
%
\tocauthor{Xiang Zhang}
\institute{University of Innsbruck, Technikerstr. 21a, Innsbruck,\\
\email{xiang.zhang@uibk.ac.at},\\ WWW home page:
\texttt{https://iis.uibk.ac.at/}
}

\maketitle              

\begin{abstract}
Constrained objects, such as doors and drawers are often complex and share a similar structure in the human environment. A robot needs to interact accurately with constrained objects to safely and successfully complete a task. Learning from Demonstration offers an appropriate path to learn the object structure of the demonstration for unknown objects for unknown tasks. There is work that extracts the kinematic model from motion. However, the gap remains when the robot faces a new object with a similar model but different contexts, e.g. size, appearance, etc. In this paper, we propose a framework that integrates all the information needed to learn a constrained motion from a depth camera into a descriptor of the constrained task. The descriptor consists of object information, grasping point model, constrained model, and reference frame model. By associating constrained learning and reference frame with the constrained object, we demonstrate that the robot can learn the book opening model and parameter of the constraints from demonstration and generalize to novel books.
\end{abstract}
\section{INTRODUCTION}

In our daily life, manipulating daily objects is complex tasks often have a certain degree of constraints for utilities and space efficiency. For instance, a door is rotating around an axle; drawers can be only pulled out in one direction; a robot needs to slide a card in a  card reader with a specific orientation to get the card properly read. These tasks share a consistent mathematical structure.

Nowadays, robots are capable of navigating in a complex environment and plan their manipulation with obstacles. In the factory, the company is exploring the possibilities of teaching a robot by human demonstration or coaching. \cite{ikeda2018flexrop} The main technical obstacle is how robots could conveniently capture,  reproduce and generalize the complex manipulation on objects especially with physical constraints
 
For a robot executing tasks, the generalization is applied in two-fold. Firstly, the robot should be able to adapt itself to different environments, e.g. obstacles. Secondly, the robot should be able to manipulate similar objects with different states (e.g. poses) and sizes. More loosely speaking, the robot should have generalizable skills for a dynamic environment.

The robot can capture and learn physical constraints from human demonstration via two means. On the one hand, a robot should learn the property of the object together with its affordance \cite{zech2017computational} with which the robot plans its behavior during execution \cite{Perez-DArpino2017}. One the other hand, researchers approach this problem by directly encoding the motion of human manipulating constrained objects. And then adapt those problems by adapting hyperparameters, namely task parameters. Traditional methods merely learn the trajectory from the demonstration with the hope that the constraints are implicitly handled by the model (\cite{Ijspeert2013}).

In one branch of research, the robot only learns the trajectory from demonstration, the generalization depends on the task parameters which is determined by the human or some hyper-parameter. In contrast, there are algorithms designed only to obtain the object model, the returned model only predicts the motion of the objects \cite{sturm2011probabilistic}, or even the motion of the objects from images \cite{Pillai2015}. Both kinds of methods are not sufficient for efficient generalization. As for generalization, given a similar object, the object model should not only detect the constraints but also return them as suitable task parameters to the robot for executing the learned policy.

In this article, we introduce a framework with a defined task descriptor to learn constrained motion from the demonstration. The learned task descriptor consists of object recognition, grasping point, task parameter detection model, constrained model, reference frame, and task parameters. The role of the reference frame is chosen where the parameters of the learned object model can be conclusively turned into executable end-effector trajectory. For instance, if a robot is learning how to open a door, the descriptor is modeled by a rotational constrained model. The rotation center is the center of the reference frame. The robot needs to learn how to identify the center of the reference frame and the parameter of the constrained motion, namely the rotation axle and radius. As a result, the motion can be generalized for any kind of doors.


This framework closes the gap between the learning object model and robot execution from the demonstration. We propose a standard procedure that integrates all necessary information, i.e. object recognition, grasping point detection, constrained motions, and reference frame, for learning and generalizing a constrained task model from the demonstration. By the integrated information from the vision system, the ambiguity of the demonstrated trajectory can be clarified by the element defined in our task descriptor.

We take color and depth frames from human demonstration and learn the descriptor of the constrained tasks.  All of the components we are proposing is for demonstrating the framework and can be replaced by more dedicated and advanced methods in the future. We specify each of the components in the following sections.
\section{Related work}

Learning from Demonstration has extensively focused on reproducing solely demonstrated trajectories \cite{Billard2008}. The constraints in the task are specified by task parameters, such as coupling terms in DMPs \cite{Zhou2017}. Other methods learn constraints implicitly by putting a restriction on the region with less variation across multiple demonstrations.

The current research of vision assisted programming by demonstration (\cite{Zhang2018}) only focus on trajectory learning, human gesture learning, object recognition, and robot navigation. Linking the vision, constraints and the resulting trajectory is still an open question.


Modeling and estimating the motion of the constrained objects attract a lot of research interest. Klingbeil \cite{Klingbeil2010} proposes a method that identifies a door handle for dealing with this particular type of articulated object. More sophisticated models are proposed describing and estimating the articulated kinematic map from observed motions \cite{sturm2011probabilistic}. Following this method, the algorithm is integrated into a vision processing algorithm so that the constrained model is automatically extracted solely from a visual recording of the manipulation articulated object \cite{Pillai2015}. Furthermore, an articulation model change detection algorithm is proposed to handle a more complex and realistic situation, such as moving a stapler and using a stapler. Huang 2012. As for faulty articulation models, Huang 2015 proposes methods that use particle filter to eliminate the wrong models and refine constrained object models.

Instead of solely rely on the perception, the constrained object model can be obtained or refined by robot interactive perception. Nekum \cite{Hausman2015} aims to extract a constrained object model in 3D space given the observation with occlusion which can be clarified by robot interaction. Other methods discover and learn the object model learning totally from scratch by interaction \cite{Katz2014}. Martin \cite{MartinMartin2014} extends the idea to the extent that the algorithm only needs visual input from the interaction.

There are also methods that cultivate robot applications. At the basic level, the robotics programmer needs to handle the constrained motion with an efficient task kinematic description and framework \cite{DeSchutter2007}. Furthermore, controllers are designed based on the articulation models which update the articulation models parameters online while interacting with objects \cite{Jain2010}. Besides adaptive controllers, Dang Proposes a framework for manipulation primitives, namely a position and a direction, where they define a task descriptors for constrained object manipulations \cite{Dang2010}. Otte integrates the articulation object research into a framework that guides the robot to proper exploration points and facilitates complex robot exploration \cite{Otte2014}. Claudia proposes a method that takes a constrained parameter and its reference frame as the basis for planning multi-step manipulation tasks.

Some of the methods rely on other sources of information. Nagahama \cite{Nagahama2018} proposes a framework that takes grasping and manipulation into consideration. With the provided constraints parameters from users, the objects from vision went through a pipeline ending up with a robot manipulation. Daniele \cite{Daniele2020} uses the visual demonstration to learn an articulated manipulation using language processing and description for model selection.

\section{Methodology}

We argue that the manipulation skill should be defined and restricted to its manipulated objects. Therefore, for every manipulation skill, the algorithm should contain the object information (namely object identification and segmentation). For manipulation, affordance, constrained model \ref{subsec::kinect} and reference frame determination are also necessary \ref{subsec::rframe}. The task descriptor of the constrained task is as follows:

\begin{itemize}
\item Identified Object: the name and location of the manipulated object

\item Grasping point and pose: grasping point for manipulation

\item Segmented objects: bounding box of the objects 

\item Constrained model: the type of constraints and its corresponding parameters.

\item Reference frame: the reference frame describing the motion, depending on the type of the constrained model. The reference frame is learned and obtained by a model that map the image of the object to a rotation center and an axle in the image.

\end{itemize}

\begin{figure}
\centering
\includegraphics[width=0.4\paperwidth]{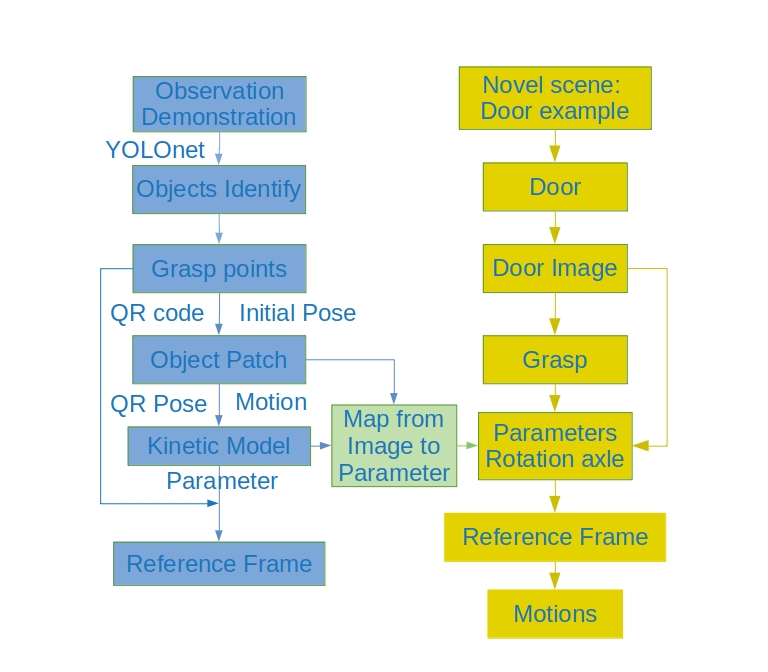}
\caption{The overall scheme}
\label{fig::scheme}
\end{figure}

We will show how the proposed framework works with simple algorithms as its elements. The details about how each of the element is constructed from demonstration and how they are recall and applied in the novel scene and objects are illustrated in fig.\ref{fig::scheme} and explained in the following sections.
\subsection{Kinematic Models Acquisition}\label{subsec::kinect}

\begin{figure}
\centering
\includegraphics[width=0.38\paperwidth]{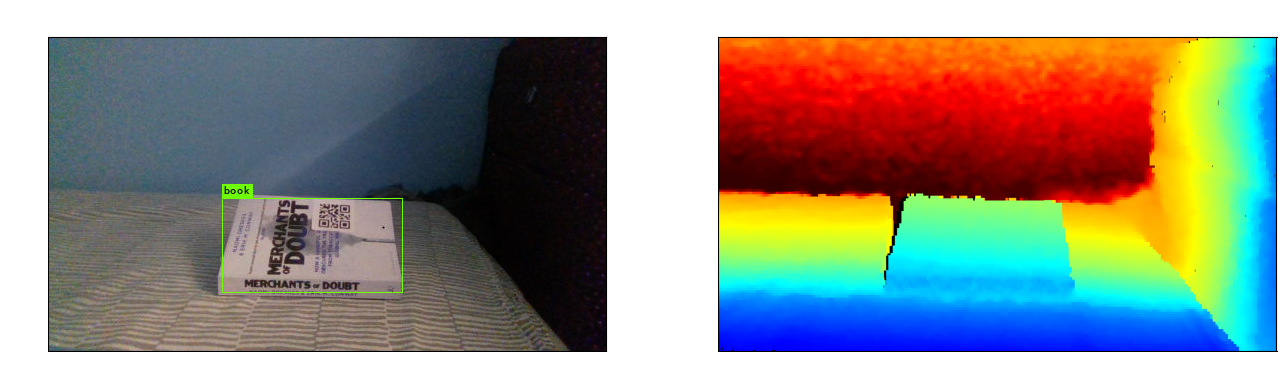}
\caption{The input image (left) and depth frame (right) from our sensor. The bounding box on the left is the result of object recognition YOLOnet-v3 \cite{redmon2017yolo9000}}
\label{fig::rawImage}
\end{figure}

Our framework depends on a depth camera. The learning procedure is shown on the left side of Fig.\ref{fig::scheme}. The input from the camera is shown in Fig. \ref{fig::rawImage}. We then use YOLO-net to identity potential objects in the scene. Then we obtain the grasping point with a QR code. The manipulated object can be then labeled with an object name. Consequently, we can segment the object out of the image with the bounding box rendered by YoLo-net.

In order to learn the model of the object from the depth image, we need to get the motion of the grasping point. Given the depth frames of the demonstration, the robot tracked the position of the grasping points in the color/depth ($p_x, p_y, d$) image which is de-projected to a point cloud ($x, y, z$) for a constrained motion. 


The reason why we choose to work within the color image instead of point cloud for rotation axle is that the point cloud is not always stable under different light conditions, whereas project pixels from 2 camera mainly depends on the camera intrinsic and relative stability to the environmental conditions. In addition, a lot of patterns on a surface cannot be detected easily from the point cloud but pretty obvious in the color image.

Subsequently, the constrained object can be described by the best kinematic graph model $\hat G$ \cite{sturm2011probabilistic} given a sequence of pose observations $\mathcal{D}$. The algorithm maximizes the posterior probability of observing the pose $\mathcal{D}$ by $\hat G$. Our application keeps the assumption of a kinematic tree.

\begin{equation}
\hat{G} = \argmax_G p(G|\mathcal{D})
\end{equation}

Under our assumption, the model is restricted to five types of articulated, namely, resolute model, prismatic model, parallel model, and rigid motion which indicate the demonstration is not a motion constrained on objects. The optimization of $G$ consists of two-part, namely selecting the model type $M$ and learning the parameter $\theta$. Each type of model has a specific form of a parameter vector that is used to construct a rotation matrix for optimization.  The optimization is done by MLESAC (maximum likelihood consensus) \cite{torr2000mlesac}.

\subsection{Reference frame Model}\label{subsec::rframe}

The reference frame is crucial to learning and generalization of the demonstrated motion. With the correct object model, the manipulation can be transferred to a novel object by tuning parameter. In order to represent the motion in a convenient and intuitive manner, we believe that the robot should be able to determine the reference frame in which the motion to be tracked by the robot is described and generated, for instance, the reference the rotation center will reduce the motion to an equation determined by radius. We only need to transform the motion back to the robot reference frame. Therefore our framework needs to learn a model that maps the initial object information of the manipulated objects to the location of the reference frame.

For instance, and the rotation center for opening a door is  Based on these structural constraints, it is possible to generalize the learned kinematic model to the other similar objects.

The reference frame is defined by a center $c$ and a vector $v$. The key component of this framework is to get the constrained object model through visions and consequently obtain the reference frame for generalization and execution. (Green Block in fig. \ref{fig::scheme}) 
The parameters in the kinematic models have the physical meaning which naturally corresponds to the shape of the observed objects. The robots are able to learn the reference frame from kinematic models from a demonstration by projecting the constrained model from section \ref{subsec::kinect} to the image, which will be used to train mapping from the raw vision into a rotation axle in the pixel coordinates. The reason why we choose the pixel frame is that the 3D pose of the object is estimated from a 2D depth image. In addition, some of the objects may not be detectable in the 3D spaces with low-cost sensors, such as a door that is nicely sealed or a thin book on the table.

We need to have a model taking the initial state of object images and grasping as input and the rotation axle as a vector in images as output. For demonstrations of the framework, from the initial state of a rotation object, we propose a simple heuristic algorithm. For objects with a revolute joint, the rotation axle $v$ is chosen either on the longer or shorter edge of the object that is further away from the grasping point. The rotation center is the perpendicular intersection point on the rotation axle through the grasping point.    

As for the prismatic constraints, the position and orientation of the grasping point and the directional vector is sufficient to place the reference frame during generalization. Similar heuristics can be defined for reference frame placement. But this is not the focus of introducing the framework.

\subsection{Manipulation}

The generalization of opening a door is demonstrated in the framework illustrated in Fig. \ref{fig::scheme}. The information is extracted in the same way as in learning. The task parameter, i.e. rotation axle and center, is first predicted from the image of the door and grasping point in the image which is de-projected to the real world for robot execution.

As mentioned, the geometric information alone is not conducive to robot task execution. In general, a motion-defined in the robot sensorimotor system can be transformed into an end-effector in the robot world coordinates. Therefore we define each type of articulation motion with respect to its identified reference frame with the following rules.

For articulated rotation, we place the reference frame at the location of the detected rotation center. The orientation of the reference frame is defined in such a way that the Y axle is aligned with the rotation axle and the z axle is aligned with the surface norm of the manipulated object. Similarly, the reference frame of the prismatic motions has centered at the grasping point and the sliding director is aligned with the frame orientation in the same way. Having the reference frame, grasping point, and the type of the constrained motion, the end-effector trajectory is calculated and transformed into a robot execution frame.

As a result, through this framework, the perceived articulated model in the demonstration generalized to novel books. The generalization heuristics can be replaced by more advanced approaches that directly taking object image as input.

\section{Experiment}

In this section, we illustrate our framework with a book-open use case. The robot learns how to open a book from the demonstration and apply this learned skill to novel books with different sizes. We do not use experiments on prismatic motions since it is simpler than rotational motion and trivial for the topic of this paper. We use Intel\textsuperscript{\textregistered} RealSense depth camera to get visual input. The camera is able to return the aligned color image, depth image, and the resulting point cloud as shown in Figure \ref{fig::rawImage}. The sampling rate is 20 Hz.

For object information, we use YOLO-net-v3 \cite{redmon2017yolo9000} to obtain the name of the manipulated objects in the scene during the demonstration. In this article, we use pre-define \textbf{QR code} to indicate the position and orientation of the grasping points since it is not the focus of our research. The encoded information in the QR code is the orientation of the grasp which is similar to the output of the grasp planning.

\subsection{Learning from Demonstration}
During the demonstration, the algorithm first recognizes the object and the grasping point from the color image. The object where the grasping point is attached is recognized by object classification as shown in Fig. \ref{fig::rawImage}. The robot then adds the object name from YOLO-net into the object information. After that, the initial object is obtained from the bounding box and saved for learning later.

\begin{figure}
\centering
\includegraphics[width=0.3\paperwidth]{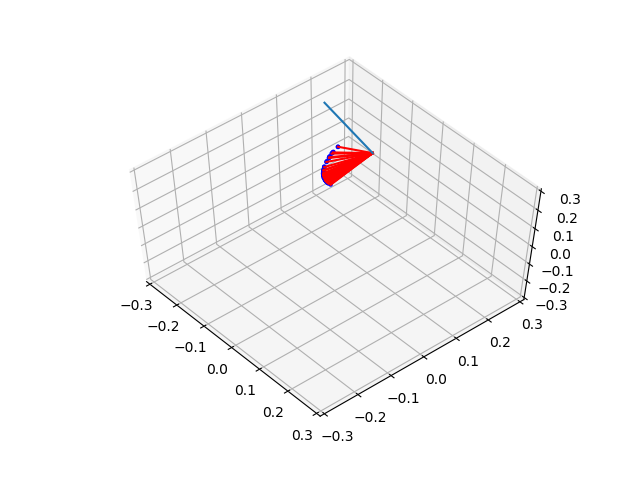}
\caption{Learning: The rotation axle obtained from the deprojected point cloud. The blue dots in the image is the blue dots each of which is connected to the rotation center with a red line and visualizes a red plane. The blue line starts from the rotation center is the rotation axle.}
\label{fig::artimodel}
\end{figure}

The trajectory of the grasping point is recorded in the 3D camera coordinates. We apply the articulation model detection algorithm described in \ref{subsec::kinect} on the recorded trajectory and obtain rotation center and axle as shown in Figure \ref{fig::artimodel}.

\begin{figure}
\centering
\includegraphics[width=0.35\paperwidth]{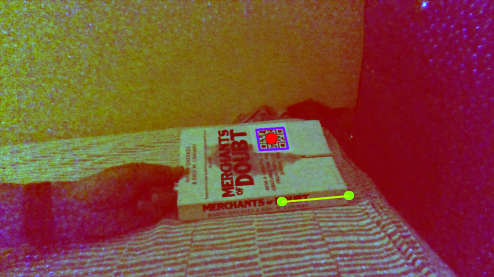}
\caption{Learning: The rotation center points and the rotation axle projected back to the initial state of the object from the point cloud. The blue dot is the rotation center and (the line between two dots are the rotation axle)}
\label{fig::pixelfeature}
\end{figure}

After that, the rotation axle is projected back to the color frame, which is shown in Fig. \ref{fig::pixelfeature}. According to the set heuristic rules in \ref{subsec::rframe}, the robot check whether the rotational axle is on the longer edge and shorter edge. And its distance to the grasping point. The robot adds them to the task descriptor for generalization. In addition, a grasping model may be learned nevertheless in our article it is replaced by QR code detection. With this information above the robot can generate the reference frame for execution.

\subsection{Generalization}

\begin{figure}
\centering
\includegraphics[width=0.35\paperwidth]{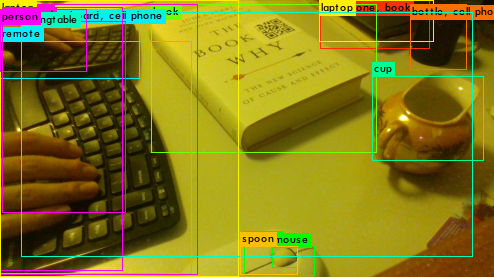}
\caption{The robot first find the book (the green bounding box) to be manipulated in a scattered environment.}
\label{fig::trajTotrack}
\end{figure}

When the robot receives command for opening a book, the robot first look for 'books' from object classification shown in Fig. \ref{fig::trajTotrack}.

\begin{figure}
\centering
\includegraphics[width=0.35\paperwidth]{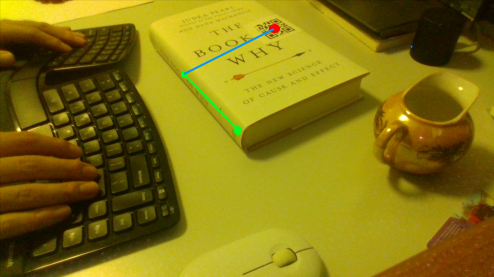}
\caption{Within the bounding box, the robot looks for the rotational axle and grasping points and consequently the radius required for execution.}
\label{fig::predictPix}
\end{figure}

Then the grasping point and object image are sent to generate the reference frame model, in this case, the robot looking for the long edge on the opposite side of the grasping point. The axle is shown as the green line in Fig. \ref{fig::predictPix}.

The predicted axle and rotation center in Fig. \ref{fig::predictPix} is de-projected back to the 3D camera coordinates. The rotation center is the origin of the reference frame and the Y axle is aligned with the rotation axle. Finally, the trajectory for the new book to be tracked is obtained by applying circle motion with the radius from grasping point to rotation axle. The result is similar to as shown in Fig. \ref{fig::artimodel} but with a different radius.








\section{CONCLUSIONS}

In this paper, we propose a framework for robot learning constrained-object manipulation, which is based on the articulation learning model. Our framework defines a descriptor of the constrained tasks that integrate all essential information required to learn and execute constrained motion from the demonstration.

The framework only on prismatic and rotational motions. The constrained motion is imposed by the manipulated objects or tasks per se, the obstacle avoidance should be applied after the constrained motion is determined. We also ignore the constraints in tool handling because the tool is designed for humans, and tools such as pliers should be provided to a robot as accessories.

The methods for the modules in our framework are simplified for the purpose of illustration. The object recognition and grasping point detection can be replaced with any method in their field \cite{zech2017computational}. In the future, we will replace the heuristic method for identifying parameters from vision with an advanced machine learning method that can be applied to more general situations.

%
\bibliographystyle{ieeetr}
\bibliography{refs}
\end{document}